\documentclass{SPAICE}
\usepackage{todonotes}
\usepackage{subcaption}

\def\authorEmail{luis.simoes@mlanalytics.ai}

\def\AuthorShort{N. Grens et al.}

\author[1, 2]{Nikki Grens}
\author[1]{Luís F. Simões\thanks{Corresponding author. E-Mail: \authorEmail}}
\author[1,3]{Kai Hou Yip}
\author[4]{Theresa Lueftinger}

\affil[1]{ML Analytics, Lisbon, Portugal}
\affil[2]{Department of Space Engineering, Delft University of Technology, Delft, The Netherlands}
\affil[3]{Department of Physics, King's College London, London, UK}
\affil[4]{European Space Agency, ESA-ESTEC, Noordwijk, The Netherlands}

\title{Traceable Spectral Inference via Influence Functions: Efficient Data Attribution and Error Proxies for the Ariel Mission}

\begin{document}

\maketitle

\begin{abstract}
Interpretability is critical for machine learning models deployed in scientific space missions such as ESA’s Ariel, where ground truth is unavailable during operations and physical plausibility must be assessed. While most explainable AI methods focus on feature attribution, this work investigates training data attribution through influence functions and introduces three key contributions for operational spectroscopy pipelines. First, influence is reformulated in terms of prediction rather than loss, enabling label-free deployment. Second, by leveraging the closed-form ridge solution of an Extreme Learning Machine, infinitesimal prediction influence is efficiently computed. Third, an influence-based conservative error proxy is derived by propagating training residuals through the influence sensitivities. Evaluated against simulated spectra, the proposed proxy correlates strongly with scale and shape-based spectral errors. Furthermore, influence functions enable the identification of the most influential samples and the approximation of the most harmful ones. Together, these results suggest that this approach can serve as an operational framework for scientific machine learning.
\end{abstract}

\section{Introduction}

Machine learning (ML) is increasingly used to extract structure from extensive datasets, including those generated by space missions \cite{carleo2019machine}. However, as model complexity grows, interpretability becomes constrained. In scientific applications where physical consistency is of great importance, an inability to understand model behaviour restricts trust in the output and complicates validation processes \cite{doshi2017towards, jia2021physics}.

The European Space Agency (ESA) M4 Ariel mission \cite{tinetti2018chemical, tinetti2021ariel} will conduct a chemical census of approximately 1000 exoplanets by analysing spectroscopic data derived during transit events. When a planet passes in front of its host star, starlight filters through the planetary atmosphere, revealing wavelength-specific molecular signatures. As a complement to the classical retrieval pipeline, ML models are being applied across different stages of exoplanet atmospheric analysis, ranging from inferring atmospheric properties from spectra \cite{Zingales_2018, Cobb_2019, Lueber_2025} to predicting transmission spectra directly from light curve data \cite{nikolaou2023lessons, simoes_2024_13885557}, which is the task addressed in this work. Since the ground truth for in-flight observations is unknown, interpretability is essential to ensure the models learn the correct physical structure.

Most explainable AI techniques focus on feature attribution using methods such as SHAP or LIME to identify which input variables influence a prediction \cite{lundberg2017unified, ribeiro2016should, guidotti2018survey}. In contrast, comparatively less attention has been directed towards training data attribution (TDA), which focuses on understanding which training samples shaped the model's behaviour in the generation of a particular prediction \cite{hammoudeh2024training}, using approaches such as influence analysis. The most intuitive approach is Leave-One-Out (LOO) retraining, where, iteratively, a single training point is removed, and the model is retrained to observe the change in the test prediction \cite{cook1982residuals}. However, this is computationally expensive even for models that train rapidly.  Influence functions, originally introduced in the context of robust statistics \cite{hampel1974influence, Cook1977} and later adapted to machine learning \cite{pmlr-v70-koh17a}, provide an efficient alternative to LOO retraining by estimating how an infinitesimal upweighting of a training point affects test predictions.

In this research, influence is formalised as the change in the predicted value of a test sample induced by a specific training instance, rather than the conventional focus on test loss, an alternative also explored in literature but for different applications \cite{pruthi2020estimating, lin2024diffusion, harilal2024influence}. This shift in definition is necessary due to the operational constraints of the Ariel mission, in which ground-truth transmission spectra are unavailable. This research specifically investigates whether this technique can serve as a framework for traceable spectral inference within a space mission pipeline by providing a transparent account of the training data driving specific predictions, whilst simultaneously evaluating the utility of these influence-based attributions as a robust proxy for error estimation in operational contexts.

\section{Methodology}
To calculate the shift in test loss without retraining, influence functions rely on a first-order approximation derived from the Hessian matrix $H$ \cite{pmlr-v70-koh17a}. This matrix represents the second-order partial derivatives of the loss function $L$ over the training dataset and describes the local curvature of the error landscape. By projecting the gradient of a training sample $z$ through the inverse Hessian $H^{-1}$ onto the loss gradient of a test sample $z_{\text{test}}$, influence analysis estimates whether upweighting that training point pushes the parameters $\hat{\theta}$ in a direction that decreases (improves) or increases (degrades) the test loss. The classical formulation is given by \cite{pmlr-v70-koh17a} with $\hat{\theta}$ the model parameters at convergence:

\begin{equation*}
I_{up, loss}(z, z_{test}) = -\nabla_{\theta} L(z_{test}, \hat{\theta})^T H_{\hat{\theta}}^{-1} \nabla_{\theta} L(z, \hat{\theta})
\end{equation*}

In this expression, $\nabla_{\theta} L(z, \hat{\theta})$ represents the gradient of the training point and $H_{\hat{\theta}}^{-1}$ the inverse Hessian. While this linear approximation is computationally efficient for standard non-linear models, it can deviate from the true effect as influence magnitudes increase and higher-order loss effects emerge.

Because ground truth is unavailable during runtime, the influence function is reformulated to quantify how each training sample shapes the model’s output response surface. This changes the quantity of interest from loss $L(z_{test}, \hat{\theta})$ to model prediction $f(x_{test}, \hat{\theta})$, a conceptual transition that does not introduce new modelling assumptions.

\subsection{Extreme Learning Machines}

The predictive model employed in this study is an Extreme Learning Machine (ELM). The ELM consists of a single hidden-layer feedforward neural network where input weights and biases are randomly initialised and fixed, and only the output weights are learned via Ridge Regression \cite{HUANG2006489, tara_elm, huang2015trends, wang2022review}. Under the squared-loss assumption, the optimisation of the output layer is convex and admits a closed-form solution. This structure is particularly advantageous for influence analysis because the objective is quadratic in the output weights, resulting in a Hessian matrix that is constant and depends only on the hidden layer activations and regularisation term. While influence in complex Deep Neural Networks must be approximated via Hessian-vector products \cite{pmlr-v70-koh17a}, the linear structure of the ELM output layer allows for a rapid and mathematically grounded calculation.

In the ELM pipeline, the output weights $\beta$ are learned by solving a Ridge Regression problem, as shown below:
\begin{equation*}
\beta = (\Phi^T \Phi + \alpha I)^{-1} \Phi^T Y
\end{equation*}
where $\Phi$ is the hidden layer output matrix, $\alpha$ is the regularisation parameter, and $I$ is the identity matrix. The matrix $(\Phi^T \Phi + \alpha I)$ corresponds to the Hessian 
$H$ of the squared-loss objective with respect to $\beta$. Under squared loss with ridge regularisation, $H$ only depends on the training feature matrix and the regularisation term. Therefore, it can be computed once and reused for all influence evaluations. 

\subsection{Prediction Influence Computation}

Due to the quadratic nature of the ELM loss objective, influence can be computed directly in closed form. For more complex architectures, approximations are required to handle the high-dimensional non-constant Hessian matrices \cite{pmlr-v70-koh17a}. The objective is to determine the influence of a training sample $z_i = (x_i, y_i)$ on the prediction $f(x_{test})$ of a test sample, and this can be done through a three-step process.
\begin{itemize}
    \item First, the training residual is calculated, where the error $r_i$ for each training point is defined as the difference between the model prediction and the ground truth.
    \item Second, the potential shift in model parameters $\theta$ is estimated by projecting the training point gradient through the inverse Hessian: $H^{-1} (\phi_i \cdot r_i)$. 
    \item Finally, the influence $I(z_i, x_{test})$ is the dot product of the test point hidden layer activations $\phi_{test}$ and the parameter change.
\end{itemize}
Note that the standard negative sign is omitted from the formulation below. This aligns the influence values with the directional shift in the prediction and the effect of removing the training sample, being purely interpretational:
\begin{equation*}
I(z_i, x_{test}) = \phi_{test}^T H^{-1} (\phi_i \cdot r_i)
\end{equation*}

\subsection{Interpretation of Influence Values}

\begin{figure*}[h]
    \centering
    
    \begin{subfigure}[t]{0.48\textwidth}
        \centering
        \includegraphics[width=\linewidth]{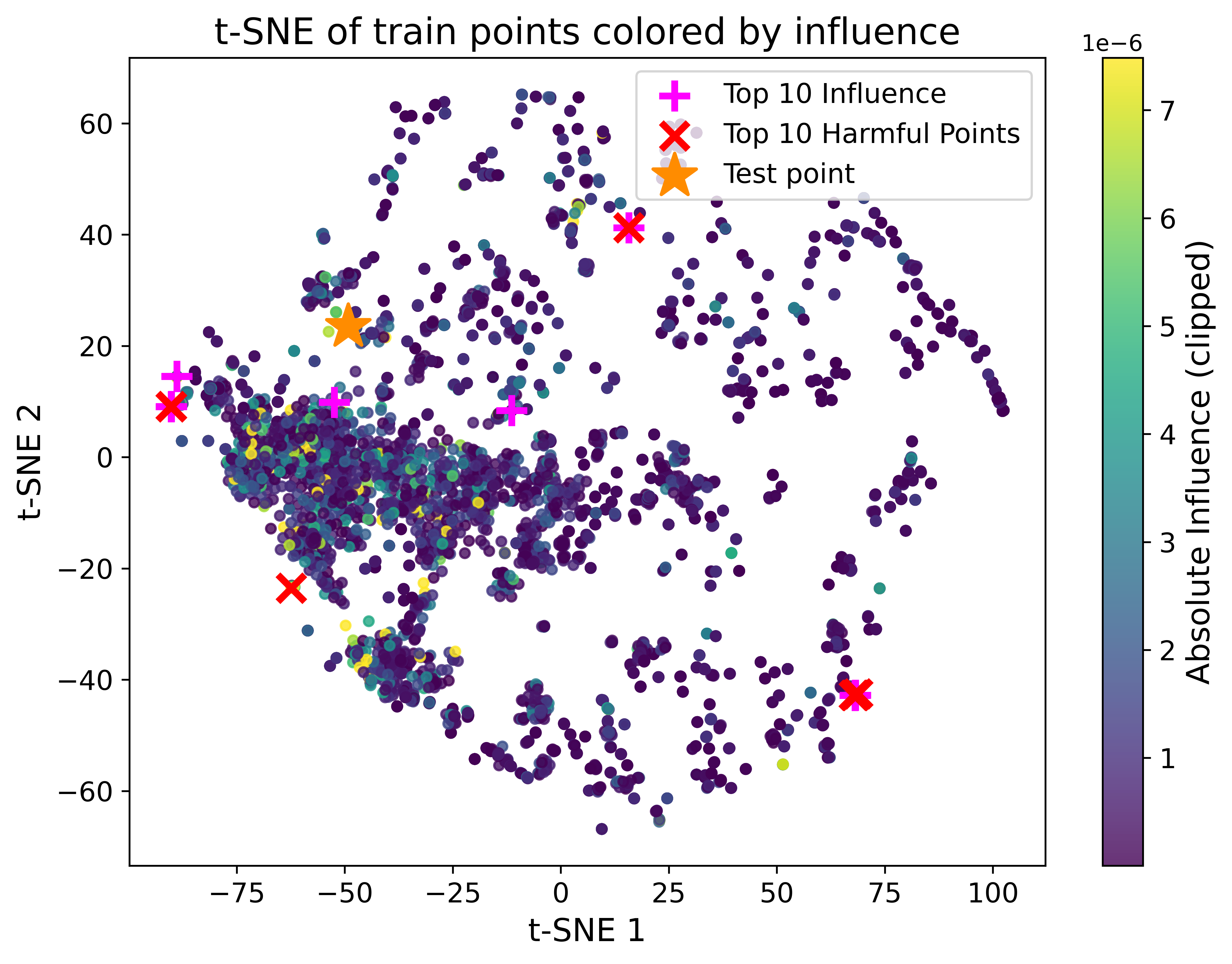}
        \label{fig:top10wl0}
    \end{subfigure}
    \hfill
    \begin{subfigure}[t]{0.48\textwidth}
        \centering
        \includegraphics[width=\linewidth]{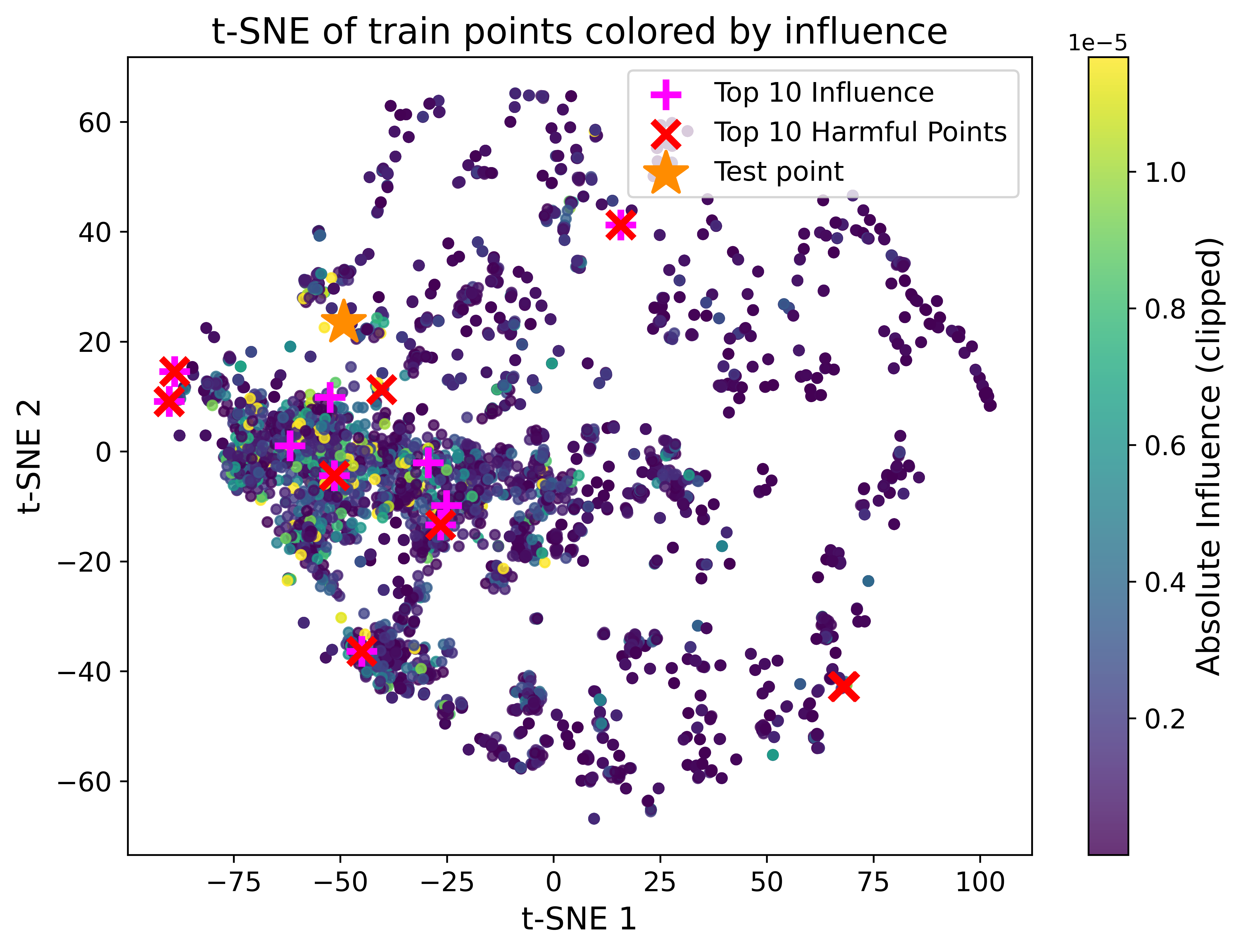}
        \label{fig:top10wl45}
    \end{subfigure}
    \vspace{-5mm}
    \caption{t-SNE embedding of training samples, coloured by absolute influence on two specific output channels: 0.55 $\mu$m (Visible, left) and 4.3 $\mu$m (Infrared, right). These channels are selected to evaluate the model's sensitivity across different physical regimes. The orange star denotes the test point under analysis. Influence values are clipped to enhance visualization, with magenta pluses and red crosses identifying the top 10 most influential and harmful samples, respectively. Notably, six samples overlap between the influential and harmful sets in both cases. Due to high proximity in the representation space, some markers appear superimposed in the figure.}
    \label{fig:top10_combined}
\end{figure*}

When influence is computed with respect to the test loss, the interpretation of the resulting values depends on the sign. A positive influence value indicates that the training sample is `harmful', as its inclusion in the training set increases the loss for a specific test prediction. Conversely, a negative value suggests the sample is `helpful'.

In this work, where influence is computed directly on the prediction rather than on the loss, the magnitude and sign show distinct dynamics. Values near zero indicate that a training sample has a negligible impact on the model's output for a given test case. Samples with large absolute influence values produce the largest local prediction shifts. A positive influence indicates that removing training point $i$ is responsible for increasing the predicted value, while a negative influence indicates that removing the training point lowers the prediction. These highly influential points, regardless of their sign, are the most critical for validation, as they include both proponents that support the model's physical consistency and opponents that may introduce bias or reflect artefacts in the synthetic training data.

\begin{figure*}[h]
    \centering
    
    \begin{subfigure}[t]{0.48\textwidth}
        \centering
        \includegraphics[width=\linewidth]{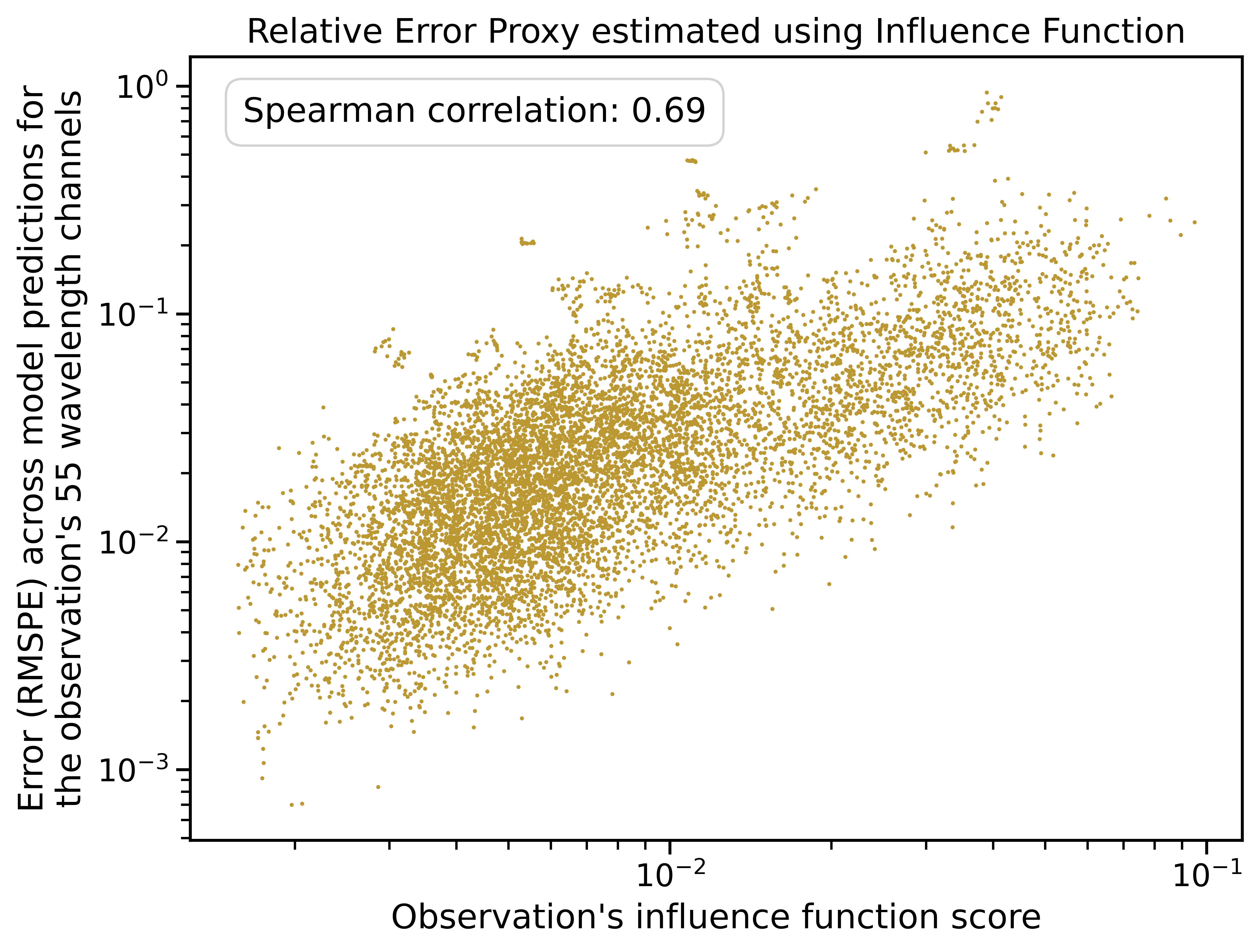}
        \label{fig:rmspe}
    \end{subfigure}
    \hfill
    \begin{subfigure}[t]{0.48\textwidth}
        \centering
        \includegraphics[width=\linewidth]{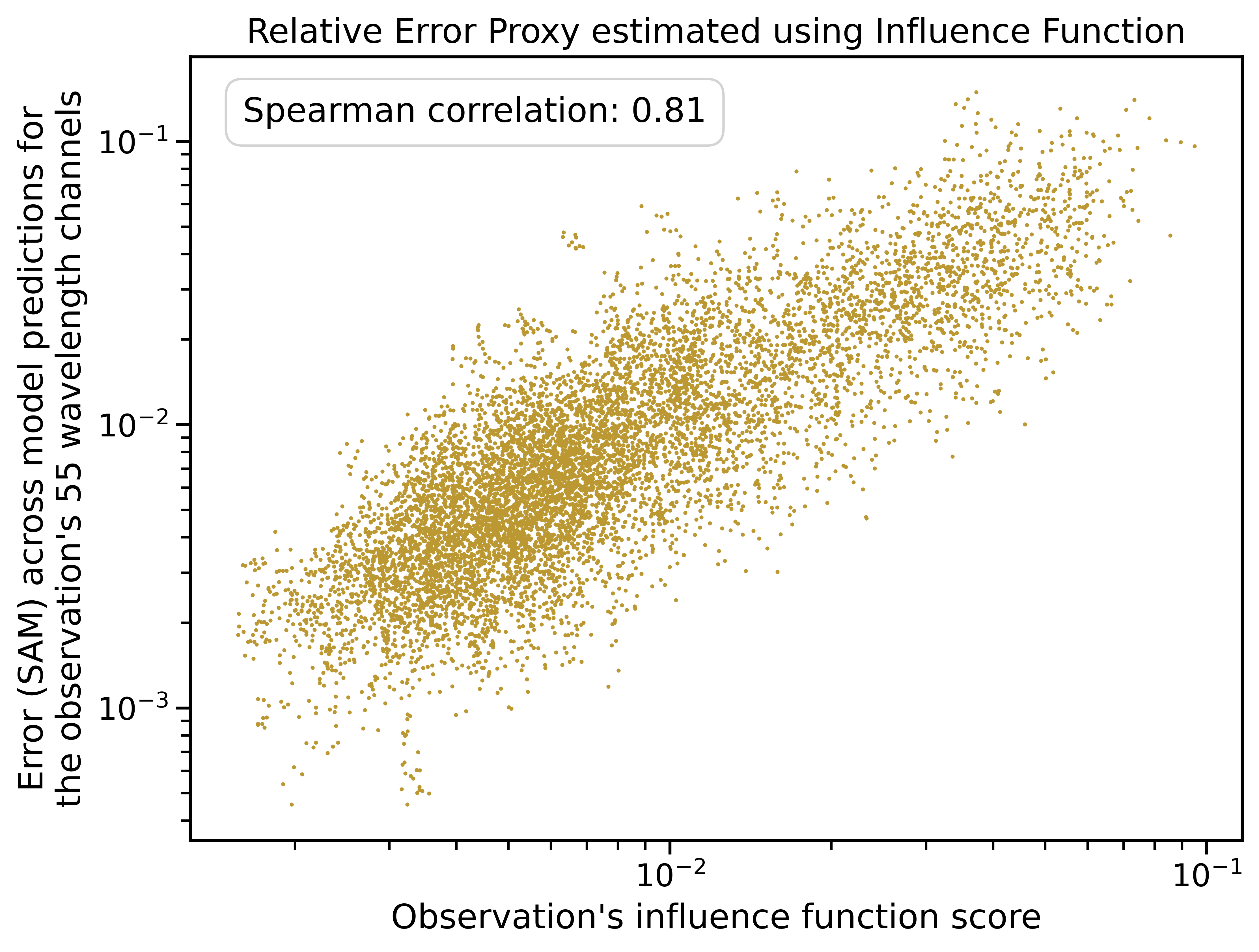}
        \label{fig:sam}
    \end{subfigure}
    \vspace{-5mm}
    \caption{Observation-level prediction error versus the influence-based proxy (log–log scale). Each point corresponds to a single test observation not seen during training. The dataset used consists of simulated data, so ground truth is available, which is used exclusively to compute the realised error. Left: RMSPE across the 55 wavelength channels shows a Spearman rank correlation of 0.69 with the influence-based error metric proxy. Right: SAM across the 55 wavelength channels shows a stronger correlation of 0.81.}
    \label{fig:correlation}
\end{figure*}

\subsection{Theoretical Extensions}
The linear structure of the ELM output layer facilitates a transition from local derivative-based sensitivity analysis to a global linear decomposition for TDA. Using the ridge solution for $\beta$, any specific test prediction $\hat{y}_{test} = \phi_{test}^T 
\beta$ can be decomposed into an additive combination of training targets. This results in a closed-form additive decomposition through the Representer Theorem \cite{representertheorem}:
\begin{equation*}
    \hat{y}_{test} = \sum_{i=1}^{N} \alpha_{i, test} y_i, \quad \text{where} \quad \alpha_{i, test} = \phi_{test}^T H^{-1} \phi_i
\end{equation*}
In this formulation, $\alpha_{i, test}$ denotes the Representer Value, representing the relative weight assigned to the training sample $i$ during the construction of the specific test prediction, which aligns with the concept of Representer Points \cite{representepoint}. Mathematically, $\alpha_{i, test}$ is equivalent to the cross-leverage between test and training points, demonstrating the close relationship between representer points and influence values.

While representer values quantify a sample's contribution to the prediction value, they do not inherently reflect the reliability of that contribution. To identify harmful points, a metric is adopted that follows the intuition of Cook's Distance \cite{Cook1977}, which identifies influential observations by combining leverage and squared residuals. In this spectral context, harmfulness is defined as ($|I(z_{i, \lambda}, x_{test}) \cdot r_{i, \lambda}|$). This identifies points where strong influence is coupled with high training error ($r_{i, \lambda}$) at a prediction output.

\section{Results}
Influence scores were obtained for an Ariel L2-L3 ML pipeline that maps transit light curves to transmission spectra \cite{nikolaou2023lessons}. This allows for predictions to be traced back to their most influential training examples in the absence of ground-truth labels.

\subsection{Experimental Setup}

The predictive model used is an ELM with a single hidden layer of 5000 neurons with sigmoid activation. Output layer weights are optimised using a regularisation $\alpha = 1.0$. The dataset is sourced from the Ariel Data Challenge 2021 \cite{nikolaou2023lessons, yip_2025_15050868}. The raw data contains 55 light curves for 55 wavelength channels, for 851 exoplanetary targets. Each planet is simulated 100 times, using 10 distinct stellar spot configurations, each with 10 photon noise instances. Preprocessing involves a multi-stage aggregation pipeline that follows the steps implemented in \cite{simoes_2024_13885557} with the following differences: the sliding window was set to nine minutes, a temporal grid of five points was used per light curve, and their spacing was defined through a dynamic grid. The resulting spectral features for the 55 channels are concatenated with relevant astrophysical metadata to form the input. These include period, transit duration, semi-major axis, inclination, eccentricity, impact parameter, star distance, star temperature, star radius, star mass, star k mag, star logg, star density, luminosity, and incident flux. After preprocessing, there are 10 samples per exoplanet, one per stellar spot configuration. All reported results were obtained through a 5-fold cross-validation (CV) process, where the data is split at a planet level to avoid data contamination.

\subsection{Evaluation Metrics}

The utility of influence functions for TDA can be evaluated through several methods. The global structure of the training data is analysed using t-distributed stochastic neighbour embedding (t-SNE) to project the high-dimensional feature space into a two-dimensional map. Each training point is coloured by its absolute influence on a specific test prediction to determine if influential samples cluster near the test point or are scattered throughout the space. Within the t-SNE projection, the most influential and harmful points are highlighted to help visualise their distribution relative to the test sample. 

An influence-based error proxy is derived by treating influence scores as first-order sensitivities that propagate training residuals to the test prediction. Under infinitesimal reweighting, the residual-weighted influence contributions are squared and aggregated into an Infinitesimal Jackknife (IJ)-inspired, non-cancelling estimate of the prediction perturbation \cite{jaeckel1972infinitesimal, Jacknifeplusinfluence, swissIJ}. For the ridge-regularised ELM, this results in a linearised approximation of the effect of infinitesimal training-sample removal on $\hat{y}_{\mathrm{test}}$. The resulting perturbation score for wavelength channel $\lambda$ is defined as $S_{\lambda}(x_{\mathrm{test}})
=
\sum_{i=1}^{N}
\left(
\phi_{\mathrm{test}}^{T}H^{-1}\phi_i\,r_{i,\lambda}
\right)^2$. To obtain an operational error proxy, a relative formulation is adopted, as a fixed absolute deviation has a larger impact for shallow transit depths than for deeper ones, shown by:
\begin{equation*}
\mathrm{Rel \ Error}(x_{\text{test}})
=
\sqrt{
\frac{1}{55}
\sum_{\lambda=1}^{55}
\frac{
\text{S}_{\lambda}(x_{test})
}{
\hat{y}_{\lambda}(x_{\text{test}})^2
}
}
\end{equation*}
For each wavelength, the score provides a conservative estimate of the induced perturbation magnitude by assuming that training-sample errors do not compensate for one another. Normalisation by the squared predicted amplitude makes the score scale-invariant, while RMS aggregation across the 55 channels produces an observation-level risk score that penalises wavelengths with large perturbations.

\section{Discussion}
The t-SNE visualisations (only illustrative and not quantitative) presented in \autoref{fig:top10_combined} demonstrate that the most influential training samples are not simply the nearest neighbours, but are spread across the training dataset. This shows that influence-based attribution captures relations beyond visual sample similarity and extends into the way samples drive model predictions of some target. The observed overlap between the most influential and most harmful samples follows directly from the definition of harmfulness. Here, `harmful' samples do not strictly guarantee an actual degradation of the test prediction, since true test labels are unobserved, but rather serve as flags for high-risk contribution. They identify training points where high cross-leverage is coupled with poor model fit, indicating that the model is relying heavily on an observation it failed to accurately capture during training. Consequently, samples with strong absolute influence dominate the harmfulness ranking, even when their individual residuals are not the largest. While the precise configuration of these points varies across wavelengths, indicating clear output dependence, the underlying structural pattern remains consistent, suggesting the model relies on broad spectral trends. Identifying the most harmful samples enables targeted data inspection, allowing astrophysicists to potentially flag high-risk spectra for removal or reweighting during model training.

To quantitatively characterise sample locations beyond t-SNE projections, scale-based Root Mean Squared Percentage Error (RMSPE) and shape-based Spectral Angle Mapper (SAM) dissimilarity metrics were calculated between the predicted spectrum of a test query $x$ and the true spectra of 50 training samples selected by minimal feature-space Euclidean distance versus maximal absolute influence scores (averaged across all 55 channels). Dissimilarities were averaged across 1710 test scenarios to provide a global comparison of feature-space proximity versus model influence. As presented in \autoref{tab:simplified_similarity}, the most influential training samples match the query's predicted spectral shape almost as well as geometric nearest neighbours, but diverge in absolute scale, resulting in a 6-fold increase in magnitude dissimilarity. This disparity demonstrates that highly influential training points do not act as mere spectral lookalikes. Instead, this observation is consistent with influence being jointly determined by cross-leverage and training-residual magnitude: training points with larger errors can exert significant influence over a test query's prediction, regardless of how closely their spectra match.

\begin{table}[htbp]
\centering
\caption{Spectral magnitude and shape dissimilarity metrics for nearest neighbours versus top influential samples.}
\label{tab:simplified_similarity}
\begin{tabular}{lcc}
\hline
Sample Subset ($k=50$) & RMSPE $\downarrow$ & SAM $\downarrow$ \\ \hline
Nearest Neighbors          & 0.1037 & 0.0309 \\
Top Influential            & 0.6278 & 0.0380 \\ \hline
\end{tabular}
\end{table}

The correlation results in \autoref{fig:correlation} quantify errors at the observation level, and build upon the framework defined by \cite{simoes_2024_13885557}. A clear monotonic relationship emerges between the influence-based error proxy and the realised prediction error for RMSPE and SAM \cite{kruse1993spectral}. The fact that a stronger Spearman correlation exists with SAM suggests that influence aligns more closely with spectral shape errors. The correlations suggest that influence-based errors can serve as a robust proxy in the absence of ground truth.

To evaluate generalisation beyond a single ELM, the proposed influence framework was applied to a more complex Gradient-Boosted \cite{hastie2009elements} Extreme Learning Machine (GB-ELM). For this architecture, influence was computed individually for the first 30 estimators of the 300-ELM ensemble to maintain computational efficiency and because, under shrinkage, subsequent models contribute less to predictions. The influences were scaled by the shrinkage factor and aggregated additively, directly matching the boosting procedure, in which each sequential model compensates for residual errors. Under this multi-model setup, the influence-based error proxy demonstrated strong alignment with the error metrics, achieving a Spearman correlation of 0.83 with SAM and 0.69 with RMSPE, slightly improving the SAM correlation while preserving the RMSPE correlation.

\begin{figure}[h]
    \centering
    \includegraphics[width=0.85\linewidth]{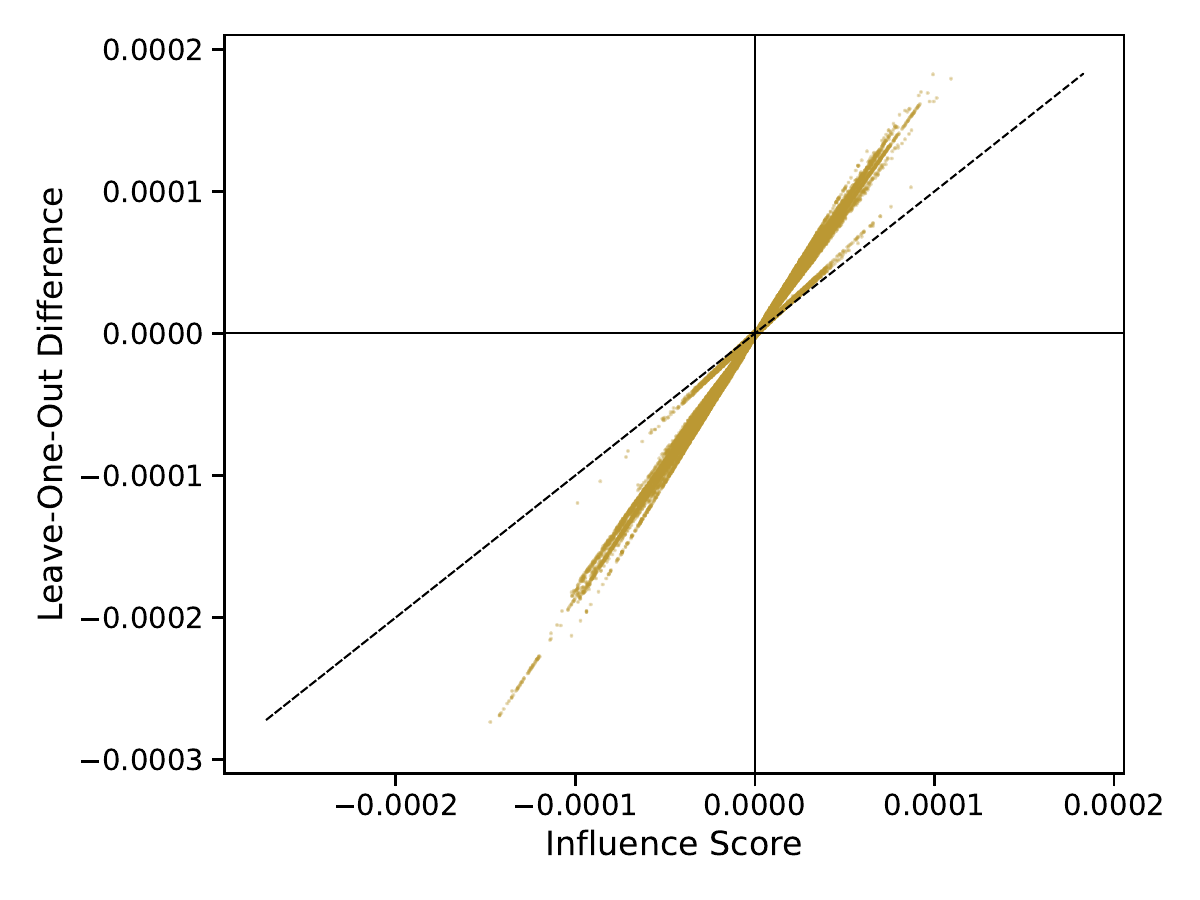}
    \caption{Impact of removing a specific training sample on a given prediction output obtained through LOO retraining and influence functions. By leveraging the linear output optimisation of the ELM, influence functions allow for an additive decomposition of the test prediction into the respective contributions of the training instances, analogous to what SHAP achieves for input features.}
    \label{fig:LOO}
\end{figure}

\autoref{fig:LOO} compares the predicted influence scores with the actual change in the model output obtained by retraining with LOO. The strong positive correlation confirms that influence functions effectively capture both the direction and relative magnitude of the retraining effect. A small subset of points aligns nearly perfectly with the diagonal, representing samples with low leverage, where influence estimates full LOO retraining almost exactly. The rest of the points display a linear correlation along a trend line steeper than the $y=x$ diagonal. Because the ELM output layer is strictly linear, this divergence does not stem from unmodelled non-linear dynamics or missing loss terms. Rather, it reflects that full sample removal in LOO retraining produces an amplification effect, governed by the leverage factor of a sample, which systematically scales the true retraining impact beyond the first-order approximation of infinitesimal sample removal.

In this study, calculating the influence matrices across all 5 CV folds, each containing 6,810 training points, required only 1 minute and 15 seconds. By comparison, performing the actual LOO retraining for just five training points per CV fold for all folds took 1 minute and 5 seconds, so a full LOO computation would have taken approximately 24 hours. Overall, the results demonstrate that influence functions provide a mechanism for efficient operational risk assessment and TDA within spectral inference.

While the exact closed-form computation of the infinitesimal prediction influence relies on the ridge structure of the ELM output layer, the overall framework is easily generalised. Specifically, the conceptual shift from test loss gradients to prediction gradients to enable label-free evaluation is architecture-agnostic and applies directly to any differentiable model. Furthermore, while the ELM provides a closed-form Hessian in a single pass, computing influence in deeper architectures has already been studied through alternative strategies, such as last-layer approximations \cite{representepoint}, iterative Hessian-vector product estimators \cite{pmlr-v70-koh17a, agarwal2017second}, or random projection methods like TRAK \cite{park2023trak}. Although deeper architectures trade away exact single-pass efficiency, the core contributions of this work, namely the label-free influence formulation, the residual-based error proxy, and the harmfulness criterion, remain conceptually applicable.

\begin{acknowledgments}
    Research funded by the ESA project ``Machine Learning and Artificial Intelligence Algorithms for Exoplanet Atmospheres Detection and Analysis'', Contract No. 4000148493/25/NL/KG.
\end{acknowledgments}

\printbibliography
\addcontentsline{toc}{section}{References}

\end{document}